\documentclass[a4paper]{cas-sc}

\usepackage[authoryear]{natbib}
\usepackage{graphicx}
\usepackage{amsmath, amssymb}
\usepackage{amsfonts}
\usepackage{algorithmic}
\usepackage{algorithm}
\usepackage{array}
\usepackage{booktabs}
\usepackage{enumitem}
\usepackage{bm}
\usepackage[caption=false,font=normalsize,labelfont=sf,textfont=sf]{subfig}
\usepackage{caption}
\usepackage[most]{tcolorbox}
\usepackage{xcolor}
\usepackage{graphicx}
\usepackage{url}
\usepackage{soul}
\usepackage{verbatim}
\usepackage{subcaption}
\usepackage{caption}
\usepackage{float}
\usepackage{booktabs}
\usepackage{multirow}
\usepackage{adjustbox}
\usepackage{tabularx}

\definecolor{promptbg}{RGB}{245,248,250}     
\definecolor{promptframe}{RGB}{0,80,130}     
\definecolor{titlegray}{RGB}{219,238,244}       
\def\tsc#1{\csdef{#1}{\textsc{\lowercase{#1}}\xspace}}
\tsc{WGM}
\tsc{QE}
\tsc{EP}
\tsc{PMS}
\tsc{BEC}
\tsc{DE}

\begin{document}
\let\WriteBookmarks\relax
\def\floatpagepagefraction{1}
\def\textpagefraction{.001}

\shorttitle{Behavior-Grounded Lane Representation Learning for Multi-Task Traffic Digital Twins}

\shortauthors{Rei Tamaru et~al.}

\title [mode = title]{Behavior-Grounded Lane Representation Learning for Multi-Task Traffic Digital Twins}                      

\author[1]{Rei Tamaru}
\ead{tamaru@wisc.edu}

\author[2]{Pei Li}
\cormark[1]
\ead{pei.li@uwyo.edu}

\author[1]{Bin Ran}
\ead{bran@wisc.edu}


\affiliation[1]{organization={Department of Civil and Environmental Engineering, University of Wisconsin-Madison},
    city={Madison},
    state={WI},
    country={USA}}

\affiliation[2]{organization={Department of Civil and Architecture Engineering and Construction Management, University of Wyoming},
    city={Laramie},
    state={WY},
    country={USA}}

\cortext[cor1]{Corresponding author}

\begin{abstract}
Traffic digital twins are powerful tools for advanced traffic management, and most systems are built on static geometric representations. However, these representations fail to capture the dynamic functional semantics required for behavior-aware reasoning, such as how a lane operates under complex traffic conditions. To address this gap, we introduce GeoLaneRep, a behavior-grounded lane representation learning framework for traffic digital twins. GeoLaneRep jointly encodes static lane geometry, observed vehicle trajectories, and operational descriptors into a shared, cross-camera semantic embedding. The encoder is trained with a joint objective combining contrastive cross-camera alignment, auxiliary role supervision, and temporal anomaly detection. Across 16 roadside cameras and 132 lanes, the learned embeddings achieve a $0.004$ lateral-rank error and an edge-role F1 of $1.000$ in zero-shot cross-camera matching, and an AUROC of $0.991$ for window-level anomaly detection. We further show that the same behavioral embeddings can condition a diffusion-based generator to synthesize lane geometries that satisfy targeted operational specifications, with $87.9\%$ overall specification accuracy across 38 lane groups. GeoLaneRep thus provides a semantic interface between roadside observations and downstream digital twin tasks, supporting cross-camera transfer, behavior-aware monitoring, and goal-directed lane synthesis. The framework is openly available at \url{https://github.com/raynbowy23/GeoLaneRep}.
\end{abstract}


\maketitle
\section{Introduction}

Digital twins have emerged as a promising system for advanced traffic management, integrating sensing, simulation, and decision-support capabilities within a cohesive virtual environment \citep{xu2023, 10992265, gu2026}. These systems continuously monitor live traffic states, evaluate control strategies, and support decision-making in the physical space. Early traffic digital twins operated primarily as descriptive platforms for visualization and state synchronization, limiting their capabilities to passive observation. This demands constructive digital twins capable of interpreting operational shifts and supporting rapid, intervention-oriented reasoning \citep{irfan2024, dasgupta2024}.

This requirement is particularly critical at the lane level, the operational layer at which traffic management interventions are defined and executed. Although contemporary digital twins replicate road geometric and vehicle trajectories at high fidelity, structural reproduction alone is insufficient for operational reasoning \citep{zhou2025, miller2021decision}. Strategic interventions, such as dynamic lane reassignment, merge coordination, turn regulation, and work-zone configuration, depend intrinsically on understanding the specific functional role each lane plays within the broader traffic ecosystem \citep{feng2025}. Therefore, a constructive digital twin must conceptualize each lane not merely as a physical boundary but as a dynamic, interpretable component whose operational utility can be analyzed and compared.

Recent advancements in representation learning offer a viable approach to bridging this semantic gap. A growing body of literature demonstrates that learned neural embeddings have been shown to extract spatiotemporal structures from high-dimensional transportation data. For instance, Trajectory Representation Learning (TRL) improves spatial similarity computations and travel-time estimation \citep{wang2024}, while graph-based architectures encode higher-order topological dependencies across macroscopic road networks \citep{mao2022, zhang2023}. Furthermore, integrated embedding strategies model multi-agent interaction ego-trajectory predictions \citep{hou2023}. These efforts demonstrate that neural representations capture rich traffic dynamics, but their focus has remained disproportionately bounded to moving agents or macro-scale network topologies, largely bypassing the lane itself as the unit of representation.

This oversight is particularly detrimental to the advancement of traffic digital twins, which require robust lane-level operational reasoning to complement geometric reconstruction. Prevailing lane representations are strictly topological or geometric, encoding centerlines, boundaries, and static connectivity. Such features are indispensable for high-definition map generation and autonomous navigation, but do not capture the functional semantics of a lane during live traffic operations \citep{mohammad2021}. Two lanes possessing identical geometric parameters may serve profoundly disparate operational purposes, ranging from uninterrupted through-movement to auxiliary flow or merge assistance. Crucially, these functional roles are not static as they emerge dynamically from the surrounding lane topology, fluctuating traffic demand, and the continuous trajectory interactions of individual vehicles \citep{poggenhans2018, naumann2023}. Consequently, current methodologies lack a representation that synthesizes static geometric structure with observed traffic behavior, thereby precluding rigorous semantic comparison across lanes and across scenes.

To address these challenges, this work proposes \textbf{GeoLaneRep}, a behavior-aware lane representation framework for traffic digital twins. GeoLaneRep jointly encodes static lane geometry, observed vehicle trajectories, and operational descriptors into a shared semantic embedding. By extending lane representation beyond static geometry, GeoLaneRep captures critical operational and functional information needed for lane-level reasoning. Experimental results on real-world data demonstrate that GeoLaneRep supports multiple downstream digital twin tasks, including zero-shot cross-camera lane matching, temporal anomaly detection, and behavior-conditioned lane geometry generation. The main contributions of this work are summarized as follows:
\begin{itemize}
    \item We introduce GeoLaneRep, a representation learning framework that produces a single shared lane embedding by jointly encoding spatial geometry, observed trajectories, and operational descriptors with three parallel encoders fused via cross-lane multi-head attention.
    
    \item We propose a joint training objective that combines contrastive cross-camera alignment, auxiliary role supervision, and temporal anomaly detection, and we show empirically that this joint formulation outperforms two-stage and single-objective variants across all evaluation metrics.
    
    \item We demonstrate that the resulting embedding supports three downstream tasks through the same encoder weights, including zero-shot cross-camera lane matching, per-window anomaly detection, and behavior-conditioned geometry generation, establishing a semantic interface between roadside observations and lane-level digital twin tasks.

    \item We evaluate GeoLaneRep using data from 16 real-world traffic cameras. Experimental results indicate strong performance in lane embedding, anomaly detection, and lane geometry generation.
\end{itemize}

\section{Related Work}

\subsection{Digital Twins for Traffic Management}
Digital twins in transportation are conceptualized as dynamic virtual counterparts of physical mobility networks, integrating sensing, communication, and computational modeling to support monitoring and decision making \citep{wang2022mobility, wang2024smart, 10992265}. Recent methodological advances have improved dynamic traffic synchronization \citep{kusic2023} and simulation-based analysis \citep{perna2025}. Nonetheless, contemporary transportation digital twins remain anchored in state alignment, visual replication, and macroscopic calibration \citep{luo2025, ocident2025}.

This paradigm presents a critical limitation: effective traffic management relies not merely on continuous observation but on interpreting the infrastructure's functional behavior as it evolves. Existing frameworks, including those designed to estimate operational measures \citep{xu2025digitaltwin} and synchronize lane geometry \citep{tamaru2025}, are predominantly observational. They maintain parity between physical and virtual states, but they lack a formal mechanism to encode lanes as actionable semantic units. Advancing toward a constructive digital twin requires a cohesive representation layer that consolidates structural geometry, traffic dynamics, and operational semantics. Addressing this methodological gap forms the primary motivation for this study.

\subsection{Semantic Understanding of Traffic Operations}
While conventional traffic models capture vehicle kinematics, they rarely yield a learned semantic representation of lane-level operations. Recent work has sought to augment traffic comprehension by deploying large vision-language models \citep{rivera2025, luo2024}. For instance, CityLLaVA \citep{duan2024} leverages roadside camera feeds for question-answering on urban scenes, whereas MAPLM \citep{cao2024} combines multi-view imagery, bird's-eye-view projections, and high-definition maps for spatial reasoning. From the ego-vehicle perspective, \citet{liao2024} fuses LiDAR and video within a VLM architecture to parse driving environments. Despite their formidable descriptive capabilities, these frameworks predominantly generate textual or symbolic outputs, such as natural language narratives, bounding boxes, and scene descriptions, which remain computationally prohibitive to integrate into the continuous mathematical control loops required for precise traffic management.

Concurrently, recent work argues that actionable traffic semantics must extend beyond visible geometry. The PAMR framework \citep{liang2026} jointly models geometric configurations and semantic traffic rules for robust navigation. Similarly, \citet{fu2024} demonstrates that explicit reasoning over lane structure enhances scene analysis, particularly in pedestrian-heavy scenarios. At a macroscopic scale, \citet{dongwei2025} incorporates road connectivity and regional proximity to refine cross-city traffic forecasting. Building upon these insights, this research pivots from descriptive, VLM-driven scene interpretation toward infrastructure-centric representation learning that embeds behavioral semantics directly into the representation.

\subsection{Representation Learning for Transportation and Lane Structure}
Representation learning is increasingly used in transportation engineering to distill latent structure from high-dimensional traffic data \citep{alahi2026}. TRL frameworks such as START \citep{jiang2023} jointly encode trajectories and road segments for self-supervised recovery, while GTR \citep{wang2025} and TRACK \citep{han2025} leverage spatio-temporal encoders and co-attentional transformers to capture intrinsic trajectory dynamics. These methods operate primarily at the network scale or in unconstrained motion settings, treating the lane as a contextual feature rather than the unit of representation.

At the lane level, several studies use representation learning to infer lane topology. \citet{li2025uni} models lanes as nodes within a unified spatial topology to predict nonlinear fluctuations in traffic flow. Complementing this, advanced vectorized mapping techniques, such as MapTR \citep{liao2022maptr}, extract structured lane networks directly from onboard sensors. Furthermore, generative approaches like CDSTE \citep{lei2024} and RoadDiff \citep{li2025} employ diffusion-based modules to infer fine-grained lane traffic states from macroscopic observations.

While these frameworks demonstrate that lane-specific topological reasoning improves predictive accuracy, their learned embeddings are optimized for narrow, task-dependent objectives. Consequently, they fail to achieve the broader mandate of representation learning: establishing a generalized methodology that uncovers underlying data structures to support versatile processing \citep{alahi2026}. To resolve this methodological shortcoming, GeoLaneRep synthesizes geometric frameworks, trajectory behaviors, and operational descriptors into a single shared embedding, transforming lanes into semantic entities that are computationally retrievable, comparable, and amenable to behavior-conditioned synthesis.

\section{Problem Setup} \label{sec:problem-setup}

\begin{figure}
    \centering
    \includegraphics[width=\linewidth]{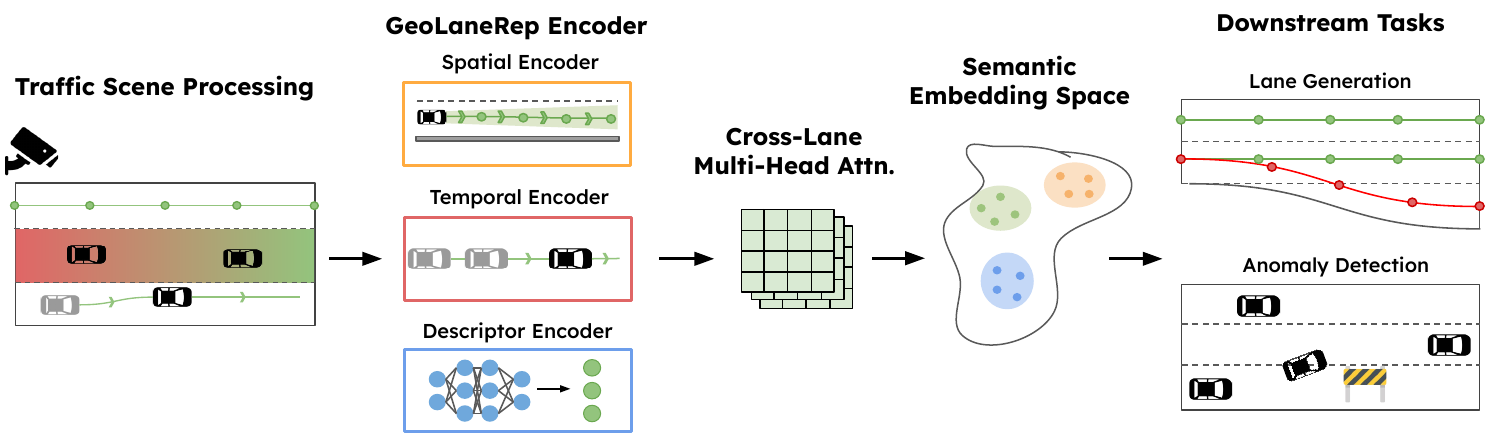}
    \caption{Overview of the GeoLaneRep pipeline. Roadside observations are converted into per-lane geometry, trajectories, and descriptors (left). Three parallel encoders, spatial, temporal, and descriptor, produce per-lane representations that are fused via cross-lane multi-head attention into a shared semantic embedding space that supports multiple downstream tasks.}
    \label{fig:geolanerep_pipeline}
\end{figure}

The GeoLaneRep framework (Figure~\ref{fig:geolanerep_pipeline}) is structured as a three-stage pipeline: \textit{observe}, \textit{encode}, and \textit{generate}. It represents a lane not as a physical shape alone but as a functional component defined jointly by its observed traffic behavior, geometric structure, and operational context. This unified representation is what lets the framework support multiple downstream tasks, zero-shot cross-camera matching, per-window anomaly detection, and behavior-conditioned lane generation, through a single shared embedding.

\subsection{Problem Formulation and Variable Definition}
\label{sec:problem-formulation}

We formulate each lane as a composite of its geometry, observed trajectories, summary traffic statistics, and structural role. For a given lane $i$,
\begin{equation}
    x_i = (g_i,\; \mathcal{T}_i,\; s_i,\; r_i),
\end{equation}
where $g_i$ is the physical lane geometry. $\mathcal{T}_i$ is the set of vehicle trajectories assigned to the lane by roadside sensing, $s_i$ holds aggregated traffic statistics, and $r_i$ encodes the lane's structural role within its lane group. 

The trajectory set is
\begin{equation}                                                              
      \mathcal{T}_i \;=\; \{\,\boldsymbol{\tau}^{(n)}_i\,\}_{n=1}^{N_i},        
      \qquad                                                                    
      \boldsymbol{\tau}^{(n)}_i \;=\; \{(x_t, y_t)\}_{t=1}^{T_n},               
\end{equation}
i.e., each lane carries $N_i$ tracklets, and each tracklet is a sequence of 2-D image-normalized waypoints over $T_n$ frames. Speed and heading are not stored per timestep; they are derived during preprocessing and enter the model only through the aggregate descriptor $s_i$. To capture short-term traffic variation without overwhelming the model with noise, we further partition each lane's trajectory stream into discrete temporal windows before feature construction. 

Beyond raw movement data, the geometric component $g_i$ provides a structural prior, formatted as a centerline polyline $g_i \in \mathbb{R}^{K \times 2}$, a sequence of $K$ waypoints in the same image-normalized coordinate frame. The descriptor $s_i$ summarizes the traffic properties derived from the assigned tracklets. The role vector $r_i$ captures interpretable structural signals such as the lane's lateral rank within its group and whether it sits at an edge of the roadway.

For a temporal window $w$, the per-window trajectory tensor is denoted as $\mathcal{T}_{i,w} \in \mathbb{R}^{N_{i,w} \times K \times 2}$, consisting of $N_{i,w}$ tracklets that have been arc-length resampled to $K$ spatial points each. The corresponding traffic descriptor $s_{i,w} \in \mathbb{R}^{4}$ records four lane-level quantities computed from those tracklets: mean speed, mean curvature, mean lateral offset from the lane centerline, and a normalized trajectory count used as a density proxy. The role vector $r_i \in \mathbb{R}^{5}$ encodes lateral rank, leftmost and rightmost edge flags, a successor flag indicating whether the lane continues into a downstream segment, and the group's lane count.

Our objective is to learn a mapping
\begin{equation}
    f \colon x_i \;\mapsto\; z_i, \qquad z_i \in \mathbb{R}^d,
\end{equation}
with $d=128$ in our implementation. The latent embedding $z_i$ is intended to preserve behavioral semantics, properly trained, it places lanes with similar operational roles close together even when their raw geometries or camera viewpoints differ substantially. This makes the representation portable across the downstream tasks introduced in the experiments.

\subsection{Input Construction and Preprocessing}
\label{sec:input-construction}

Each lane instance is constructed from three primary information streams: static geometry, dynamic vehicle movement, and interpretable descriptors. The geometric stream provides the physical backbone of the lane. The trajectory stream captures how drivers actually interact with that structure, isolating realized behavior from idealized lane shape. The descriptor stream summarizes operational signals derived from those interactions, aggregate kinematics ($s_i$), and the lane's structural role within its group ($r_i$).

Constructing these inputs requires assigning trajectories to candidate lanes through spatial association with the known geometry within each lane group. We define a \textit{lane group} as a set of adjacent lanes that share the same cross-section and local flow context (e.g., the four lanes of a single direction at a single intersection approach). The group provides the basis for determining lateral rank, edge status, and the neighbor-aware features used by the cross-lane attention. 

Observations are then organized into discrete temporal windows, preserving short-term traffic fluctuations while simultaneously filtering extraneous noise. Geometry is normalized into a lane group's shared image-normalized coordinate frame, and trajectory statistics are aggregated per window. Prior to encoding, the anchor geometry $g_i$ is broadcast across the time dimension of the per-window stream, and the per-window dynamic statistics are concatenated with the static role vector to form a single unified descriptor:
\begin{equation}
    x^{\mathrm{stat}}_{i,w} = [\,s_{i,w} \parallel r_i\,] \in \mathbb{R}^{9}.
\end{equation}

The preprocessing pipeline, therefore, yields aligned, multi-modal inputs in which each lane is represented as a coherent semantic unit -- geometry, behavior, and structural context bound together -- rather than as an isolated geometric line on a map.

\section{GeoLaneRep Encoder} \label{sec:geolanerep_encoder}

\begin{figure}
    \centering
    \includegraphics[width=\linewidth]{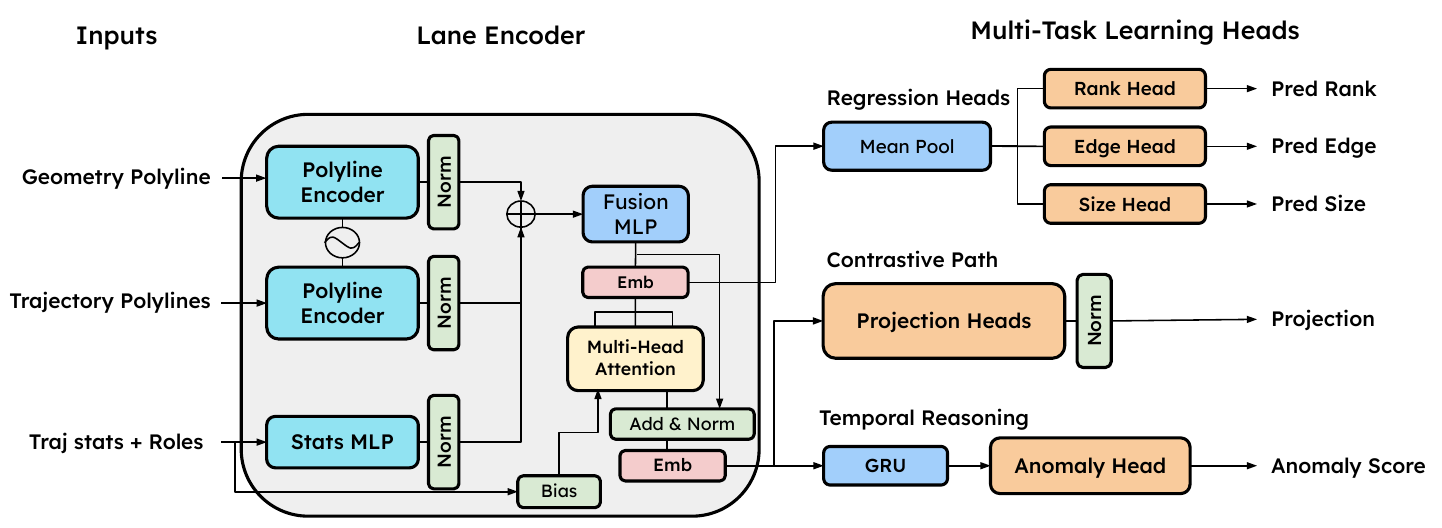}
    \caption{GeoLaneRep encoder architecture. The static lane geometry, the per-window set of assigned trajectories, and the fused stats--role descriptor pass through three parallel encoders that each emit a 64-dim embedding. The three embeddings are concatenated and passed through a fusion MLP to produce the per-window lane embedding, then averaged over valid temporal windows to obtain a per-lane embedding.}
    \label{fig:geolanerep_encoder}
\end{figure}

Figure~\ref{fig:geolanerep_encoder} presents the architecture of the GeoLaneRep encoder. The preprocessed lane inputs are encoded into a shared embedding space that captures lane-level behavioral semantics. To make latent similarity reflect operational similarity, the encoder is trained with structurally mined contrastive supervision across disparate camera views, supplemented by auxiliary role regression and temporal anomaly objectives that prevent representational collapse during long training runs.

\subsection{Input Streams and Fusion} \label{sec:encoder_fusion}

For lane $i$ and temporal window $w$, the encoder consumes three inputs: (i) a static lane geometry polyline $g_i \in \mathbb{R}^{K\times 2}$, (ii) the set of trajectory polylines $\mathcal{T}_{i,w} \in \mathbb{R}^{N_{i,w}\times K\times 2}$ together with a validity mask $m_{i,w}\in\{0,1\}^{N_{i,w}}$ that flags which slots correspond to real tracklets, and (iii) a fused descriptor vector $x^{\mathrm{stat}}_{i,w}$ formed by concatenating the four trajectory statistics $s_{i,w}\in\mathbb{R}^{4}$ with the five-dim structural role descriptor $r_i\in\mathbb{R}^{5}$.

The architecture processes these inputs through a geometry stream, a trajectory stream, and a descriptor stream, each emitting a 64-dim embedding. The three embeddings are then fused into a single per-window lane representation.

\paragraph{Geometry Stream.}
The geometry branch maps the static lane shape into a structural embedding,
\begin{equation}
    f_i^{g} = f_{\mathrm{geom}}(g_i) \in \mathbb{R}^{64}.
\label{eq:geom-stream}
\end{equation}
$f_{\mathrm{geom}}$ first applies a per-waypoint linear projection $\mathbb{R}^{2}\to \mathbb{R}^{64}$, adds a fixed sinusoidal positional encoding $\mathrm{PE}_K$ to preserve spatial ordering, and runs the resulting sequence through a 2-layer Transformer encoder. The token sequence is mean-pooled across the $K$ waypoitns to a single vector, after which a per-stream Batch Normalization is applied externally.

\paragraph{Trajectory Stream.}
The trajectory branch models how vehicles traverse the lane,
\begin{equation}
    f_{i,w}^{x} = f_{\mathrm{traj}}(\mathcal{T}_{i,w}, m_{i,w}) \in \mathbb{R}^{64}.
\label{eq:traj-stream}
\end{equation}
$f_{\mathrm{traj}}$ shares the geometry stream's per-polyline encoding scheme: each tracklet's $K$ waypoints are projected positionally encoded, and Transformer-encoded to a 64-dim vector. The resulting per-tracklet embeddings are subsequently mean-pooled across the $N_{i,w}$ slots with validity mask $m_{i,w}$ used to zero out padded slots and to divide by the true valid count, so padding does not bias the mean. Per-stream Bach Normalization is applied externally.

\paragraph{Descriptor Stream.}
The third branch processes the interpretable descriptor, which carries operational signals, aggregate kinematics, edge membership, and lateral ordering, not easily recovered from raw polylines. The descriptor passes through a two-layer MLP,
\begin{equation}
    f_{i,w}^{s} = f_{\mathrm{desc}}\bigl([s_{i,w} \parallel r_i]\bigr) \in \mathbb{R}^{64}.
\label{eq:stats-stream}
\end{equation}
where $f_{\mathrm{desc}}$ has the form of a two-layer network mapping from $\mathbb{R}^{9} \to \mathbb{R}^{64}$, with per-stream Batch Normalization and GELU activations so that all three streams share the same output normalization scheme.

\paragraph{Fusion.}
The three branch outputs are concatenated to form an intermediate representation $c_{i,w} = [\,f_i^g \parallel f_{i,w}^x \parallel f_{i,w}^s\,] \in \mathbb{R}^{192}$, which is projected to the final per-window embedding,
\begin{equation}
    z_{i,w} = f_{\mathrm{fuse}}(c_{i,w}) \in \mathbb{R}^{128}.
\label{eq:per-window-emb}
\end{equation}
$f_{\mathrm{fuse}}$ expands the 192-dim concatenation to 256-dim intermediate space via a linear projection, applies GELU activation and dropout, and projects down to the final $\mathbb{R}^{128}$ embedding.

A global, per-lane embedding is obtained by averaging across $W_i$ valid temporal windows of lane $i$,
\begin{equation}
    \bar z_i = \frac{1}{W_i}\sum_{w=1}^{W_i} z_{i,w},
\label{eq:global-emb}
\end{equation}
This fusion design lets every lane be summarized by a single $\mathbb{R}^{128}$ vector while preserving per-window granularity for the temporal anomaly head, which consumes the per-window sequence $\{z_{i,w}\}_{w=1}^{W_i}$ directly.

\subsection{Cross-Lane Attention} \label{sec:cross-lane-attn}

In multi-lane road segments, a lane's operational meaning is inherently relative. A passing lane is defined by its speed relative to adjacent traffic; an edge lane derives its role from the physical boundary. To capture these group-relative semantics, GeoLaneRep optionally routes the per-lane embeddings through a cross-lane attention module before downstream use.

Lanes that share a common group identifier are packed together for joint processing. For every ordered lane pair $(i,j)$ within a group, three relative features are combined:

\begin{equation}
    \phi_{ij} = \bigl[ \Delta_{\mathrm{lateral}}(i,j),\; \Delta_{\mathrm{speed}}(i,j),\; \rho_{\mathrm{density}}(i,j) \bigr],
\label{eq:rel-features}
\end{equation}
where $\Delta_{\mathrm{lateral}}(i,j) = s^{\mathrm{lat}}_i - s^{\mathrm{lat}}_j$ and $\Delta_{\mathrm{speed}}(i,j) = s^{\mathrm{spd}}_i - s^{\mathrm{spd}}_j$ are signed differences between the per-lane mean lateral offset and mean speed of the assigned trajectories, while $\rho_{\mathrm{density}}(i,j) = \frac{s^{\mathrm{cnt}}_i}{s^{\mathrm{cnt}}_j + \varepsilon}$ is the ratio of their normalized trajectory counts, clamped to $[-10,10]$ to prevent numerical explosion.

The pairwise feature stack $\Phi \in \mathbb{R}^{|\mathcal{G}|\times|\mathcal{G}|\times 3}$ is projected linearly to a per-head bias of dimension equal to the number of attention heads, and added directly to the scaled dot-product attention scores. A multi-head self-attention block with a residual connection and layer normalization then yields the final group-aware embedding,
\begin{equation}
    \tilde z_i = \mathrm{LayerNorm}\bigl( \bar z_i + f_{\mathrm{MHA}}(\bar z_i, \mathcal{Z}_{\mathrm{group}(i)}, \phi_{i,\cdot}) \bigr) \in \mathbb{R}^{128}.
\label{eq:cross-lane-attn}
\end{equation}
where $\mathcal{Z}_{\mathrm{group}(i)}$ denotes the set of per-lane embeddings for all lanes in the group containing lane $i$, and $\phi_{i,\cdot}$ collects the relative features between lane $i$ and every group member. Padded slots produced during batched group packing are masked out of both the attention scores and the output, so they exert no influence on the attended representation.

\subsection{Multi-Task Learning}
\label{sec:multitask}

The encoder is trained under a multi-task regimen in which several heads supervise distinct facets of the learned representation. Rather than acting as isolated task-specific outputs, these heads shape the shared embedding space toward structural consistency, semantic interpretability, and temporal sensitivity.

\subsubsection{Contrastive Alignment}
\label{sec:proj-head}

To enforce cross-camera structural alignment, the group-aware embedding $\tilde z_i$ is mapped through projection head $f_{\mathrm{proj}}$ before being $\ell_2$-normalized:
\begin{equation}
    p_i = f_{\mathrm{proj}}(\tilde z_i), \qquad \hat p_i = \frac{p_i}{\|p_i\|_2}.
\label{eq:proj-head}
\end{equation}
$f_{\mathrm{proj}}$ operates as a two-layer multi-layer perceptron mapping $\mathbb{R}^{128} \to \mathbb{R}^{64}$. The trailing BatchNorm decorrelates batch statistics before $\ell_2$ normalization, which lets cosine similarity reduce cleanly to the dot product $\hat p_i^{\top}\hat p_j$.

The normalized projection $\hat p_i$ is then optimized with the standard InfoNCE objective \citep{oord2018}, following the contrastive representation learning paradigm \citep{ting2020, xinlei2020}, with positive pairs mined structurally across cameras rather than through input augmentation. This pulls structurally similar lanes from distinct cameras together in the embedding space while pushing apart pairs that disagree in role. Following standard representation learning practice, this projection head is discarded at inference, so downstream tasks operate directly on the encoder embeddings.

\subsubsection{Role Supervision} \label{sec:role-regression}

A purely contrastive objective leaves the embedding free to ignore structurally meaningful attributes, an outcome typically called \emph{representation collapse}, where the encoder finds a shortcut that minimizes the contrastive loss without preserving useful geometry. Three auxiliary heads provide direct supervision against this failure mode. They attach to the pre-attention per-lane embedding $\bar z_i$ rather than to the post-attention $\tilde z_i$, so the supervision targets a per-lane signal that is not yet contaminated by within-group mixing:

\begin{itemize}
    \item \textbf{Lateral rank head} $f_{\mathrm{rank}} \colon \mathbb{R}^{128} \to \mathbb{R}$, a two-layer MLP through a 32-dim hidden layer, predicts the lane's normalized lateral position within the lane group.
    \item \textbf{Edge flag head} $f_{\mathrm{edge}} \colon \mathbb{R}^{128} \to \mathbb{R}^{2}$, a single linear projection, outputs logits for the leftmost and rightmost flags.
    \item \textbf{Group size head} $f_{\mathrm{size}} \colon \mathbb{R}^{128} \to \mathbb{R}$, a linear projection, estimates the group's normalized lane count.
\end{itemize}

All three heads emit raw logits and are trained with BCE-with-logits against bounded $[0,1]$ targets, which provides stronger gradients at extremes than MSE on a sigmoid-squashed prediction would.

The combined role loss is
\begin{equation}
    \mathcal{L}_{\mathrm{role}} =\mathcal{L}_{\mathrm{rank}}+\mathcal{L}_{\mathrm{edge}}+0.5\,\mathcal{L}_{\mathrm{size}}.
    \label{eq:role-loss}
\end{equation}

When cross-lane attention is active, an additional group-rank consistency term enforces that, within each group, the predicted ranks form a monotonic and roughly uniformly spaced sequence:
\begin{equation}
    \mathcal{L}_{\mathrm{group}} = \frac{1}{G} \sum_{g=1}^{G} \left\| \mathrm{sort}\bigl(\sigma(\hat r_g)\bigr) - \mathrm{linspace}(0,1,n_g) \right\|_2^2,
\label{eq:group-consistency}
\end{equation}
where $\hat r_g$ collects the predicted rank logits for the $n_g$ lanes in group $g$ and $\sigma$ is the sigmoid function.

\subsubsection{Joint Training Objective} \label{sec:joint-training}

In the contrastive stage, the encoder loss combines the InfoNCE term with the role losses under a three-phase epoch schedule:
\begin{equation}
    \mathcal{L}_{\mathrm{E}}(e) = w_{\mathrm{ctr}}(e)\,\mathcal{L}_{\mathrm{ctr}} + w_{\mathrm{role}}(e)\,\mathcal{L}_{\mathrm{role}}.
\label{eq:encoder-loss}
\end{equation}

The schedule (with $e/E$ the training fraction) is
\begin{equation}
    (w_{\mathrm{ctr}}, w_{\mathrm{role}})(e)
    = \begin{cases}
    (0.3,\;2.0) & e/E < 0.3\quad\text{(role-dominant warm-up)}\\
    (1.0,\;1.0) & 0.3 \le e/E < 0.7\quad\text{(balanced)}\\
    (2.0,\;0.5) & e/E \ge 0.7\quad\text{(contrastive fine-tune).} 
    \end{cases}
\label{eq:phase-schedule}
\end{equation}
The schedule prevents the contrastive and role objectives from colliding early in training: role supervision establishes an interpretable per-lane signal first, contrastive alignment then pulls cross-camera matches together, and a final contrastive-heavy phase sharpens retrieval quality.

The temporal branch learns from synthetic anomaly injection. A fraction of valid training windows are corrupted by simulated speed reductions, trajectory drop-outs, or lateral deviations. The anomaly head is supervised by a validity-weighted binary cross-entropy loss,
\begin{equation}
    \mathcal{L}_{\mathrm{temp}} = \frac{
        \sum_{i,w}v_{i,w}\,\mathrm{BCE}\!\bigl(a_{i,w}, y_{i,w}\bigr)
    }{
        \sum_{i,w}v_{i,w}
    },
    \label{eq:temporal-loss}
\end{equation}
where $v_{i,w}\in\{0,1\}$ is the window validity indicator and $y_{i,w}\in\{0,1\}$ is the ground-truth anomaly label, so padded windows contribute neither numerator nor denominator, and the loss remains comparable across batches with different valid-window counts.

Under the joint training configuration, the encoder remains trainable through both streams, and the total loss is
\begin{equation}
    \mathcal{L} = \alpha\,\mathcal{L}_{\mathrm{temp}} + \beta\,\mathcal{L}_{\mathrm{E}},
    \label{eq:joint-loss}
\end{equation}
where $\alpha, \beta$ control the relative influence of the temporal and structural objective. Empirically, this joint formulation is what makes the encoder useful for both spatial retrieval and downstream temporal reasoning rather than excelling at one at the expense of the other (Sec.~\ref{sec:loco-eval}).

\subsection{Geometry Dropout for Zero-Shot Transfer} \label{sec:geom-dropout}

Cross-camera generalization is a defining requirement for GeoLaneRep. A lane observed by a previously unseen camera may lack reliable geometric annotation, yet it must still be matched against a reference lane bank. To prevent the encoder from overfitting to static structural geometry, we apply geometry dropout during training.

With probability $p_{\mathrm{drop}}$, the geometry-stream embedding $f_i^g$ is zeroed out. This forces the network to derive lane semantics from the trajectory and descriptor streams alone. To preserve expected activation magnitudes, non-dropped embeddings are rescaled by $1/(1-p_{\mathrm{drop}})$, the standard inverted-dropout correction. At inference, geometry can be leveraged for established reference lanes that carry annotations while safely omitted for new query lanes. This technique enables trajectory-driven zero-shot retrieval without retraining.

\section{Downstream Tasks} \label{sec:downstream}

\begin{figure}
    \centering
    \includegraphics[width=\linewidth]{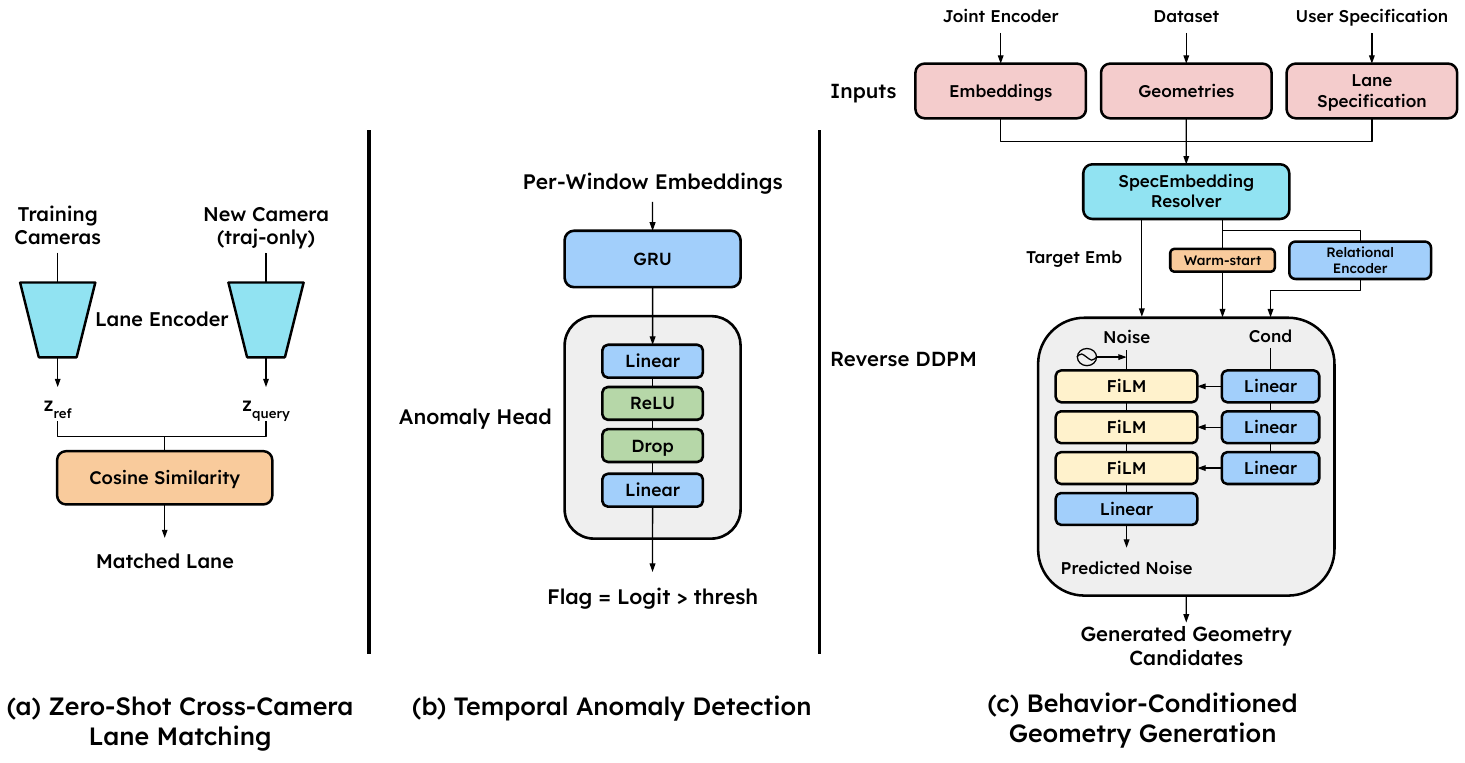}
    \caption{Downstream tasks built on the GeoLaneRep encoder. (a) Zero-shot cross-camera lane matching. (b) Temporal anomaly detection. (c) Behavior-conditioned geometry generation.}
    \label{fig:geolanerep_downstream}
\end{figure}

Following the training phase, the GeoLaneRep encoder facilitates three downstream tasks, as Figure~\ref{fig:geolanerep_downstream} shows. These operate directly on the learned embeddings rather than on raw geometry or trajectories. This section details three primary applications: zero-shot cross-camera lane matching, temporal anomaly detection, and behavior-conditioned geometry generation. All three reuse the encoder weights from the contrastive/joint training described in Sec.~\ref{sec:multitask}. Only the generation task adds a separately trained head.

\subsection{Zero-Shot Cross-Camera Lane Matching} \label{sec:matching}

Given a query lane from a previously unseen camera, the matching task retrieves the most similar lane from a reference bank constructed over the training cameras. Both query and reference lanes are encoded with the trained encoder, and the match is the reference whose embedding has the highest cosine similarity to the query:

\begin{equation}
    \mathrm{match}(q) = \arg\max_{r \in \mathcal{R}}
    \frac{\bar z_q^{\top}\bar z_r}{\|\bar z_q\|\,\|\bar z_r\|}.
    \label{eq:match}
\end{equation}

This setup tests whether the encoder captures viewpoint-invariant structural semantics rather than scene-specific shortcuts. In deployment, reference lanes are typically encoded with all modalities available, while a query lane may be encoded under restricted conditions -- most importantly, with the geometry stream dropped (Sec.~\ref{sec:geom-dropout}) when the new camera has no attention. The matched reference's structural attributes (lateral rank, edge flags, group size) transfer to the query lane, providing the operational labels needed by the rest of the pipeline.

\subsection{Temporal Anomaly Detection} \label{sec:temporal_downstream}

To facilitate continuous temporal reasoning, the framework processes each lane as a sequence of per-window embeddings:

\begin{equation}
    Z_i = [z_{i,1}, z_{i,2}, \dots, z_{i,W_i}].
    \label{eq:temporal-seq-downstream}
\end{equation}
and feeds the sequence through GRU to obtain a hidden state at each window:
\begin{equation}
    h_i^{(w)} = \mathrm{GRU}\!\bigl(z_{i,w}, h_i^{(w-1)}\bigr).
    \label{eq:gru-downstream}
\end{equation}

A small MLP anomaly head maps each hidden state to a scalar logit $a_{i,w} \in \mathbb{R}$, supervised at training time by the validity-weighted BCE loss $\mathcal{L}_{\mathrm{temp}}$ defined in Eq.~\ref{eq:temporal-loss} against synthetically injected anomalies. 

At inference, the logits are passed through a sigmoid to obtain per-window anomaly probabilities, and a threshold (chosen by Youden's J on the validation split) converts them to binary anomaly flags. Because the GRU consumes the full per-window sequence, the decision at window $w$ has access to the lane's preceding behavioral context rather than only its current trajectory snapshot.

\subsection{Behavior-Conditioned Geometry Generation} \label{sec:generation}

To connect the learned representation to digital twin interventions, the framework incorporates a behavior-conditioned geometry generator. Given a target embedding $z^\star \in \mathbb{R}^{128}$ that specifies a desired operational profile, supplied by the user or retrieved from a reference lane with the desired role, the generator synthesizes a candidate centerline $\hat g$ whose behavioral semantics, when re-encoded, are close to $z^{\star}$.

The module is a denoising diffusion probabilistic model (DDPM) \citep{ho2020} with a linear $\beta$ schedule of length $T_{\mathrm{diff}}$. Rather than initializing from pure Gaussian noise, the generative process applies a warm-start forward diffusion from a canonicalized anchor geometry $g_{\mathrm{anchor}}$ flattened into $w_s = \mathrm{vec}(g_{\mathrm{anchor}}) \in \mathbb{R}^{32}$ ($K=16$ waypoints $\times$ 2 coordinates) by running the forward diffusion only up to an intermediate step $t_0$:
\begin{equation}
    x_{t_0} = \sqrt{\bar{\alpha}_{t_0}}\, w_s + \sqrt{1-\bar{\alpha}_{t_0}}\,\epsilon, \qquad \epsilon \sim \mathcal{N}(0,I),
    \label{eq:warm-start}
\end{equation}
where $\bar{\alpha}_t = \prod_{s=1}^{t}(1-\beta_s)$.

The reverse denoising process is conditioned on the target embedding $z^\star$ via Feature-wise Linear Modulation (FiLM) \citep{perez2018}. At each denoiser layer $l$, two linear projections of the conditioning vector produce a scale and shift,
\begin{align}
    \gamma^l &= f_{\mathrm{scale}}(z^\star), \label{eq:film-scale} \\
    \beta^l  &= f_{\mathrm{shift}}(z^\star), \label{eq:film-shift}
\end{align}
which modulate the layer's hidden state after a linear transform and LayerNorm:
\begin{equation}
    h_l = \mathrm{GELU}\!\Bigl( \mathrm{LayerNorm}\bigl(f_{\mathrm{hidden}}(h_{l-1})\bigr) \cdot (1+\gamma^l) + \beta^l \Bigr).
    \label{eq:film-layer}
\end{equation}

The denoiser also consumes a sinusoidal timestep embedding concatenated with the noisy geometry; from this it predicts the noise to remove at the current step.

Iterating the FiLM-conditioned denoiser from $t_0$ down to $0$ yields a candidate geometry $\hat g$ that has been continuously nudged toward the target embedding's behavioral semantics. The module produces an ensemble of candidates per specification, which are scored and ranked by re-encoding $\hat g$ and measuring its similarity to $z^{\star}$ in the encoder space.

\section{Experiments and Results}

This section evaluates the GeoLaneRep framework across both encoder performance and downstream task execution. The evaluation encompasses lane-centered dataset construction, controlled preprocessing, and a series of targeted experiments. These experiments are explicitly designed to test representation quality, temporal consistency, and generation utility.

\subsection{Experiment Setup}

\begin{figure}
    \centering
    \includegraphics[width=0.9\linewidth]{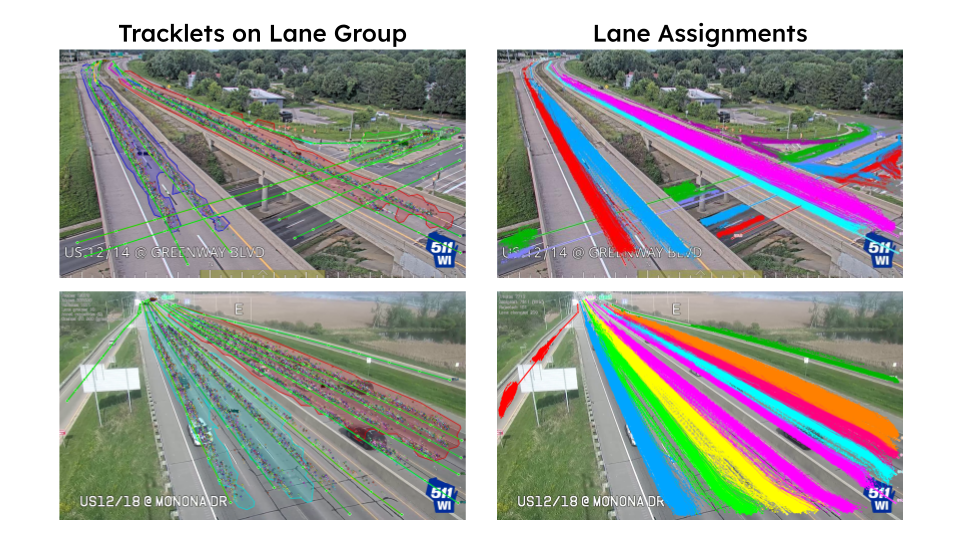}
    \caption{(Left) Tracklets extraction for each lane group depicted by transparent boundaries. (Right) Tracklets assignments to each annotated lanelet. Within a lane group, each color marks a distinct lane.}
    \label{fig:preprocessing}
\end{figure}

The dataset is collected and prepared to expose variations in lane geometry, traffic behavior, and cross-site operating conditions. It aggregates feeds from 16 roadside cameras served by 511~Wisconsin, covering 132 lanes organized into 38 lane groups and a total of 104{,}415 vehicle trajectories. The structural composition reflects diverse operational functionalities: 38 leftmost (28.8\%), 38 rightmost (28.8\%), and 38 interior (28.8\%) lanes, with the remaining 18 lanes (13.6\%) functioning as merges. All 16 cameras contribute to both training and cross-camera evaluation.

To prepare the raw video feeds, the framework uses a deterministic preprocessing pipeline. Raw vehicle kinematics are first extracted using YOLOv11n \citep{khanam2024} with a confidence threshold of $\ge 0.25$, coupled with persistent tracking. Tracklets surviving a 5-frame minimum are spatially normalized to a $[0, 1]$ coordinate space. As illustrated in Figure~\ref{fig:preprocessing}, these dynamic tracklets are then assigned to manually annotated static lane boundaries. Assignment computes the mean point-to-polyline distance between each tracklet and every candidate lane, binds the tracklet to the nearest lane within a 60-pixel threshold, and discards tracks that fail the threshold as out-of-bound tracking errors.

Crucially, this pipeline converts raw kinematic data into a structured lane-level representation. Rather than treating each vehicle in isolation, inputs are aggregated so that every sample describes holistic lane behavior, including temporal fluctuations and the lane's operational role (leftmost, rightmost, interior, or merge). Each lane sample is summarized by four aggregate trajectory statistics: mean speed, mean curvature, mean lateral offset from the lane centerline, and a normalized trajectory count used as a density proxy, computed over the tracklets assigned to that lane. 

All static geometries and dynamic trajectories are arc-length resampled to $K=16$ points via linear interpolation. To enable contrastive learning, positive pairs are strictly mined across different cameras. A candidate pair is accepted only when the two lanes satisfy three structural alignment criteria: lateral rank difference ($|\Delta| < 0.15$), identical edge-type flags (leftmost/rightmost), and role cosine similarity ($\ge 0.8$). The final collated batches supply the encoder with standardized tensors: static geometry $(B, K, 2)$, dynamic trajectories $(B, T_{\max}, K, 2)$ with a boolean validity mask $(B, T_{\max})$ that prevents zero-padded slots from biasing the mean-pooled trajectory embedding, and the 9-dim fused stats vector $(B, 9)$ concatenating four trajectory statistics with the five-dim structural role descriptor.

\subsection{Leave-One-Camera-Out Cross-Camera Evaluation}
\label{sec:loco-eval}

Cross-camera generalization is evaluated using a leave-one-camera-out (LOCO) protocol. For any given camera $c$, all lanes from the remaining cameras $\mathcal{C}\setminus\{c\}$ are encoded to form a reference bank. Lanes from the held-out camera then serve as queries, each matched to its nearest reference by cosine similarity in the learned embedding space. This procedure is repeated with every camera held out in turn. This configuration tests whether the representation captures lane identity in a manner that transfers across unseen viewpoints and novel roadway layouts.

The evaluation reports four complementary metrics. \textit{Mean cosine similarity} measures overall nearest-neighbor affinity. However, this metric alone can be misleading, because a representation that collapses every lane into a narrow region of feature space will also produce uniformly high similarities. Greater analytical weight is therefore placed on \textit{lateral rank difference}, $|r_q-r_{\mathrm{ref}}|$, which verifies that the matched reference shares the query's relative lateral position rather than an arbitrary nearby embedding. This evaluation also tracks \textit{edge F1}, which checks preservation of leftmost and rightmost boundary roles, alongside \textit{anomaly accuracy}, measured on the same LOCO splits for methods that expose a temporal detection module.

\subsubsection{Comparison Across Models}
\begin{table*}[t]
\centering
\caption{Comparison across cross-camera matching and anomaly detection baselines under the LOCO protocol. Lower lateral-rank difference is better; higher edge-role F1 and anomaly accuracy are better. "---" denotes metrics not applicable to a given method. The \emph{Is Generalize} column summarizes whether the method is designed to transfer to unseen cameras at test time: \emph{yes} for methods trained without per-site labels, \emph{no} for supervised or oracle baselines, \emph{partial} for \texttt{traj-stats}, which uses no learned representation but requires the target site's own trajectories at query time.}
\label{tab:model_comparison}
\resizebox{\textwidth}{!}{
\begin{tabular}{l l c c c c c}
    \toprule
    Method & Supervision & Match sim. $\uparrow$ & Lat. diff $\downarrow$ & Edge F1 $\uparrow$ & Anomaly $\uparrow$ & Is Generalize \\
    \midrule
    traj-stats         & none                & 0.991 & 0.398 & 0.356 & 0.875 & partial \\
    stats-oracle (GT)      & role labels required& 0.990 & 0.018 & 1.000 & --- & no \\
    per-camera-sup     & per-camera labels   & ---   & 0.227 / 0.476 & 0.481 / 0.182 & --- & no \\
    SVM (One-Class)    & none                & ---   & ---   & ---   & 0.776 & yes \\
    LSTM (traj-stats)  & none                & ---   & ---   & ---   & 0.823 & yes \\
    \midrule
    GeoLaneRep (Ours) & & & & & & \\
    \midrule
    two-stage (frozen) & none                & ---   & ---   & ---   & 0.821 & yes \\
    geometry-only      & none                & \textbf{1.000} & 0.448 & 0.262 & ---   & yes \\
    trajectory-only    & none                & 0.996 & 0.387 & 0.358 & ---   & yes \\
    contrastive        & none                & 0.932 & 0.013 & 0.994 & ---   & yes \\
    joint       & none                & 0.962 & \textbf{0.004} & \textbf{1.000} & \textbf{0.979} & yes \\
    \bottomrule
\end{tabular}
}
\end{table*}

Table~\ref{tab:model_comparison} benchmarks the proposed methodology against representative non-learning baselines, supervised references, and anomaly-detection alternatives. The compared models operate as follows: 
\begin{itemize}
    \item \textbf{traj-stats} uses a minimal set of aggregate trajectory statistics per lane and performs nearest-neighbor matching without learned embeddings.
    \item \textbf{stats-oracle} serves as the matching upper bound. It reads ground-truth lane-role annotations directly as input, so its high matching scores are not achievable at deployment time. It has no temporal module, hence no anomaly result.
    \item \textbf{per-camera-sup} trains supervised lane-role predictors independently for each camera. Reporting both the within-camera validation and the held-out-camera test scores isolates whether camera-specific supervision transfers to unseen views.
    \item \textbf{contrastive} uses the static encoder trained purely for cross-camera alignment, with no temporal modeling. 
    \item \textbf{two-stage (frozen)} establishes cross-camera alignment first, freezes the encoder weights, and subsequently trains a GRU-based anomaly detector on those fixed embeddings.
    \item \textbf{One-Class SVM} and \textbf{LSTM (traj-stats)} are anomaly-focused baselines. They indicate whether joint representation learning is required for temporal detection.
    \item \textbf{geometry-only} and \textbf{trajectory-only} are ablated architecture variants that isolate static shape cues and dynamic motion cues, respectively.
    \item \textbf{joint} constitutes the full proposed model. The encoder and temporal anomaly module are optimized simultaneously end-to-end. 
\end{itemize}

The core finding is that high raw matching similarity does not imply meaningful cross-camera correspondence. The geometry-only, trajectory-only, and traj-stats baselines all yield extremely high nearest-neighbor similarity scores ($0.991$--$1.000$), yet their lateral-rank errors remain large ($0.387$--$0.448$), and their edge-role F1 scores remain low ($0.262$--$0.358$). The three representations evidently place lanes into narrow clusters that admit high cosine similarity to the nearest neighbor without preserving the lane's functional role. 

Applying the contrastive encoder sharpens cross-camera discriminability: lateral-rank error drops to $0.013$ and edge F1 rises to $0.994$, at the cost of a slightly lower raw similarity ($0.932$). The full joint model refines this further, reaching a lateral-rank error of $0.004$ and an edge F1 of $1.000$, while simultaneously securing the highest anomaly detection accuracy ($0.979$).

These results lead to two conclusions. First, cross-camera lane correspondence requires the combination of geometric structure, trajectory observation, and contrastive alignment. No single modality alone is sufficient. Second, joint temporal-and contrastive training strengthens the embedding for both spatial retrieval and downstream temporal reasoning, rather than sacrificing one for the other.

\subsubsection{Training Behavior: Joint vs.\ Two-Stage Optimization}

\begin{figure*}
    \centering
    \begin{minipage}[t]{0.38\textwidth}
        \centering
        \includegraphics[width=\linewidth]{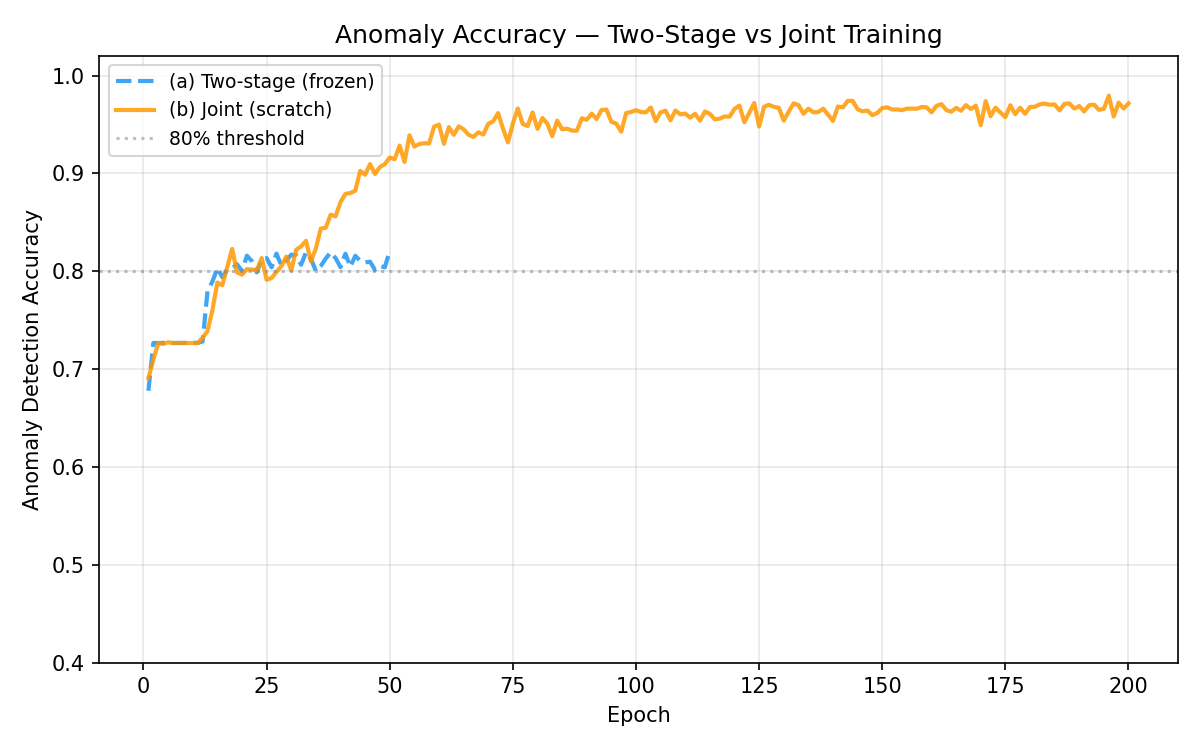}
    \end{minipage}
    \hfill
    \begin{minipage}[t]{0.58\textwidth}
        \centering
        \includegraphics[width=\linewidth]{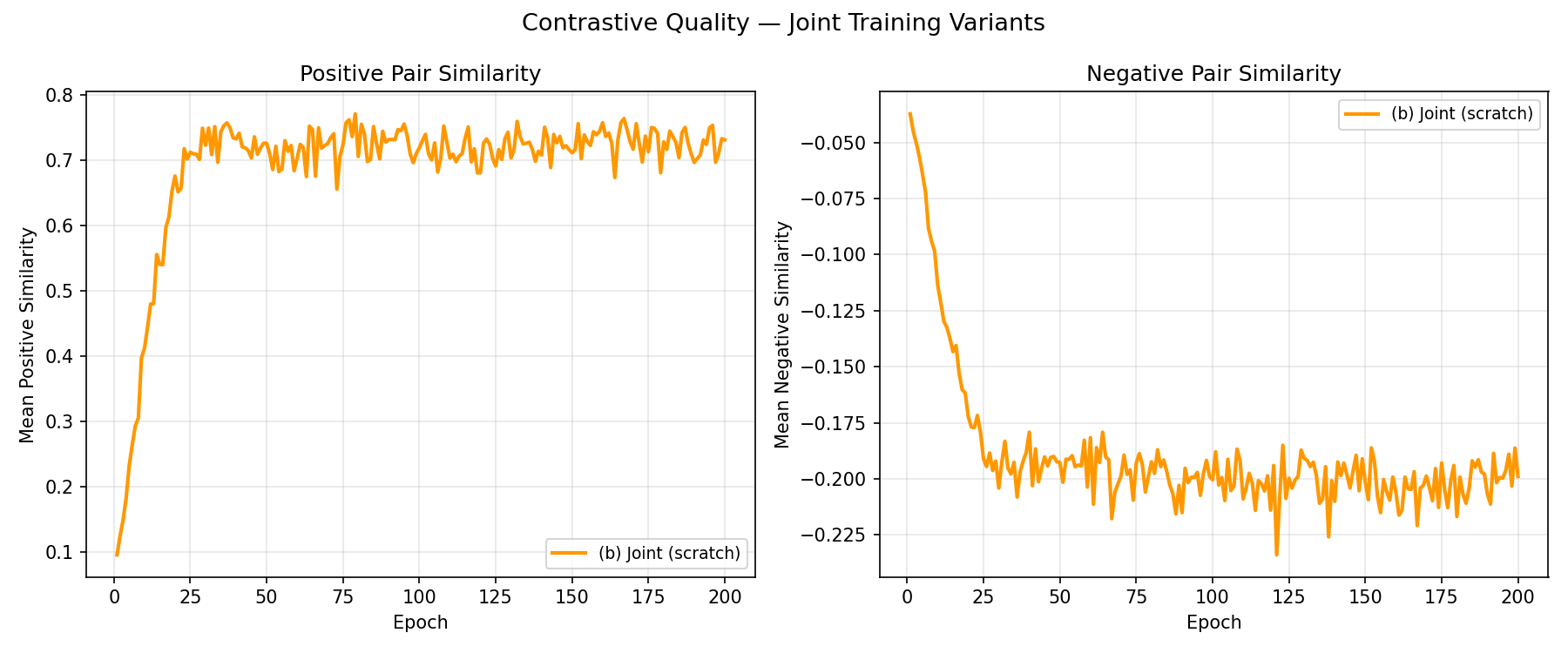}
    \end{minipage}
    \caption{Training behavior of the proposed model. Left: anomaly detection accuracy for two-stage training with a frozen encoder versus end-to-end joint training. Joint optimization continues improving well beyond the plateau reached by the frozen-encoder variant. Right: contrastive quality during joint training, showing rapid convergence of positive-pair similarity and stable separation from negative pairs.}
    \label{fig:training_behavior}
\end{figure*}

Analyzing the training strategies exposes a clear gap between end-to-end joint optimization and the two-stage approach. Figure~\ref{fig:training_behavior} (Left) shows the frozen-encoder variant plateauing near $81\%$ anomaly accuracy by epoch $\sim 25$ and making negligible further progress over the remaining epochs of that run. The joint model crosses the same $80\%$ threshold at epoch $\sim 25$ and continues improving, stabilizing around $97\%$ by epoch $\sim 75$. This sustained divergence indicates that anomaly detection capability cannot be merely ``bolted on'' to a fixed representation. The encoder itself must adapt to the temporal objective.

Figure~\ref{fig:training_behavior} (Right) confirms that the contrastive structure remains intact during joint training. Mean positive-pair similarity rises from $\sim 0.10$ to $\sim 0.72$ within the initial 25 epochs. Simultaneously, mean negative-pair similarity drops to $\sim -0.20$ over the same interval. Both curves remain stable across the full 200-epoch run, with variance at the scale of mini-batch noise rather than systematic drift. End-to-end optimization, therefore, heightens temporal sensitivity without triggering representation collapse (positives and negatives remaining well-separated) or eroding cross-camera alignment.

\subsection{Representation Quality}

\begin{figure*}
    \centering

    \begin{minipage}[t]{0.30\textwidth}
        \centering
        \includegraphics[width=\linewidth]{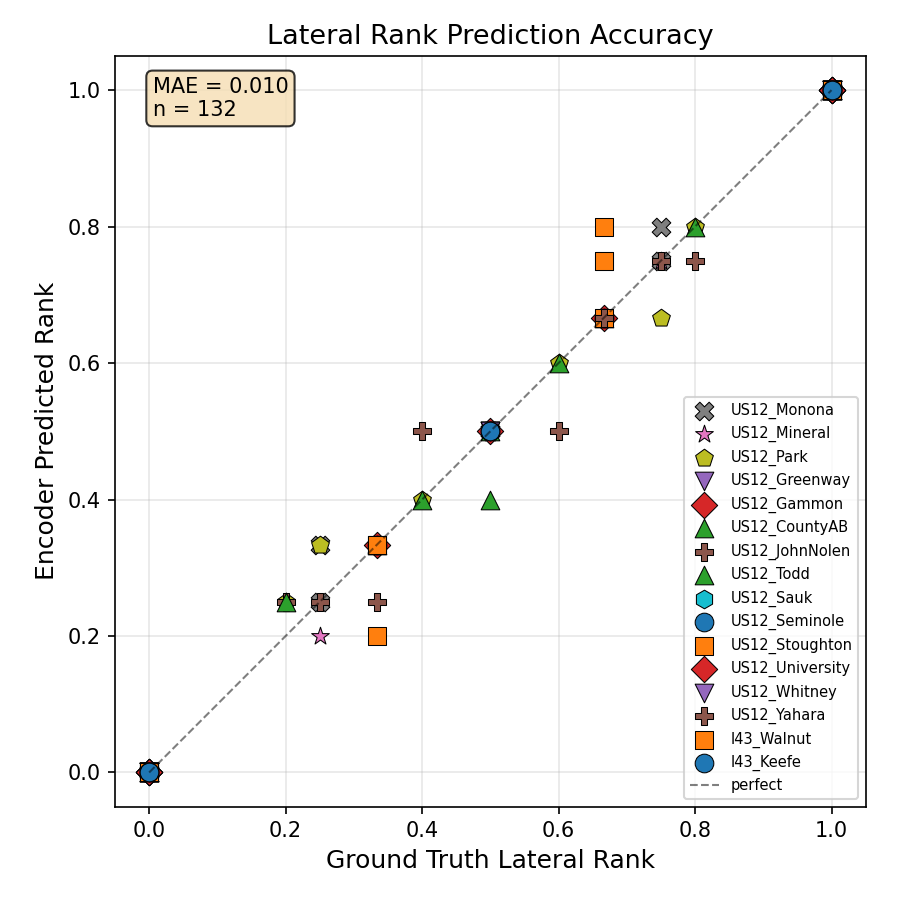}
        \vspace{1mm}
        \textbf{(a) Lateral rank alignment}
    \end{minipage}
    \hfill
    \begin{minipage}[t]{0.66\textwidth}
        \centering
        \includegraphics[width=\linewidth]{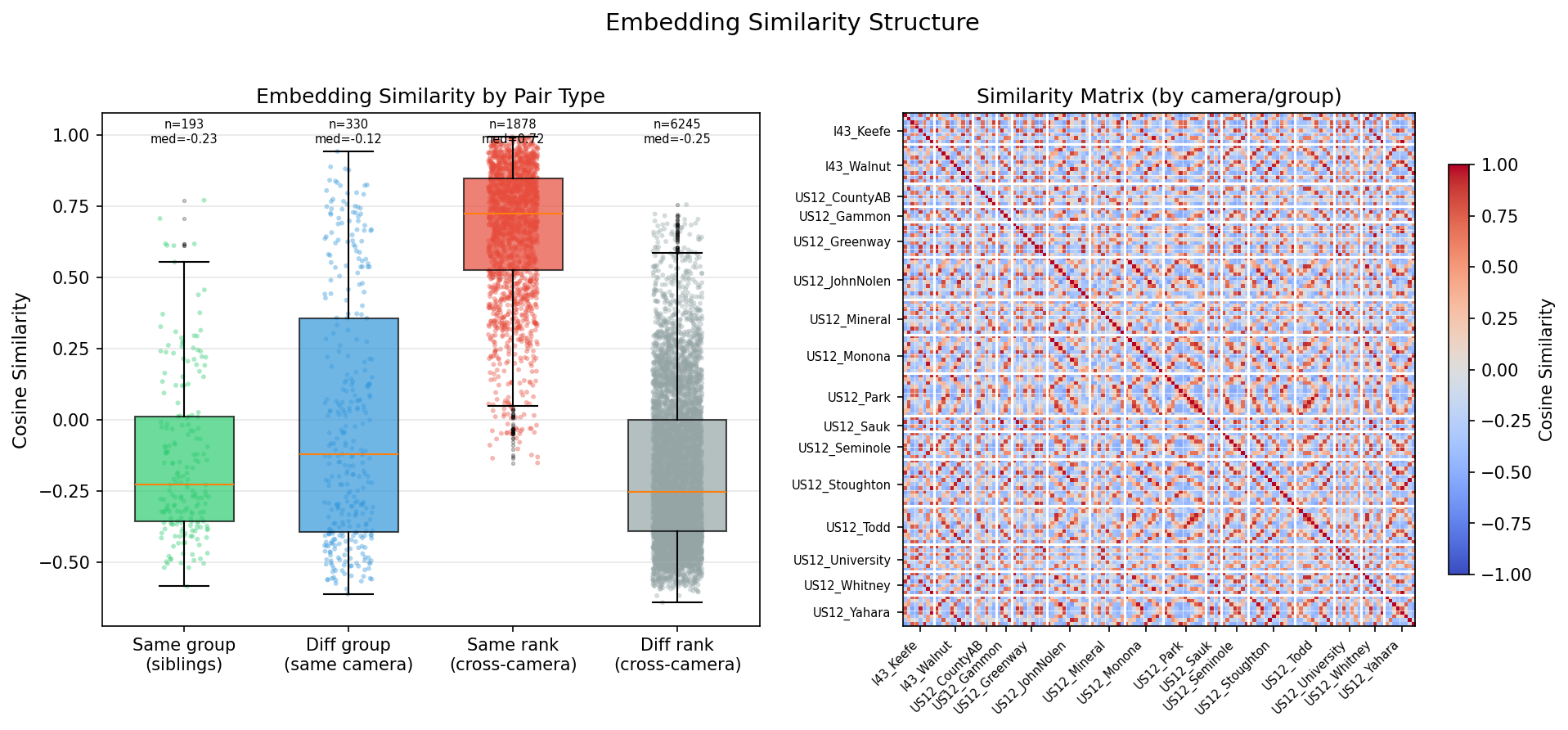}
        \vspace{1mm}
        \textbf{(b) Embedding similarity structure}
    \end{minipage}

    \caption{
    Cross-camera lane embedding evaluation.
    (a) Cross-camera lateral rank alignment across 132 lanes from 16 cameras. Each point plots a query lane's ground-truth lateral rank against the ground-truth rank of its embedding-nearest neighbor retrieved from a different camera.
    (b) Embedding similarity structure. Left: cosine similarity of projection pairs grouped into four categories. Right: per-lane cosine similarity matrix, organized by camera and group.
    }
    \label{fig:embedding_results}
\end{figure*}

\subsubsection{Lateral Rank Prediction}
The encoder produces an embedding space in which cross-camera nearest-neighbor retrieval preserves lateral position: for each of the 132 lanes, the embedding-nearest lane drawn from a different camera shares the query's ground truth lateral rank with mean absolute error $0.010$ (Figure~\ref{fig:embedding_results}(a)). Matches align with the diagonal across the full $0\!\to\!1$ rank spectrum. This confirms the encoder captures the relative lateral ordering of lanes independent of the raw camera viewpoint under which a lane was observed.

\subsubsection{Embedding Space Structure}
Probing the learned embedding space reveals a distinct geometry driven by the contrastive training objective (Figure~\ref{fig:embedding_results}(b)). Lanes sharing identical lateral ranks across disparate cameras generate a median cosine similarity of $+0.72$ ($n=1{,}878$ pairs). This reflects strong cross-scene generalization at the semantic level. Conversely, sibling lanes situated within the exact same physical group, which share spatial proximity but execute different roles, are pushed apart to a median similarity of $-0.23$ ($n=193$), essentially matching the cross-camera diff-rank baseline. The embedding, therefore, does not use camera or group co-occurrence as a similarity shortcut; rank identity dominates. Different groups within the same camera sit at an intermediate $-0.12$ ($n=330$), indicating only a small residual camera bias. The per-camera similarity matrix (right panel) corroborates this structure; block-diagonal intensity captures within-camera within-group identity, while sparse off-diagonal bright cells mark the cross-camera same-rank matches that drive the $+0.72$ distribution.

\subsection{Temporal Consistency and Anomaly Sensitivity}
This evaluation isolates whether the learned embedding maintains temporal stability under normal operation while reacting dynamically to operational disruptions. The embedding is assessed as an anomaly signal across three dimensions: absolute detection accuracy, robustness to temporal aggregation scale, and structural localization.

\begin{figure}
    \centering
    \includegraphics[width=0.8\linewidth]{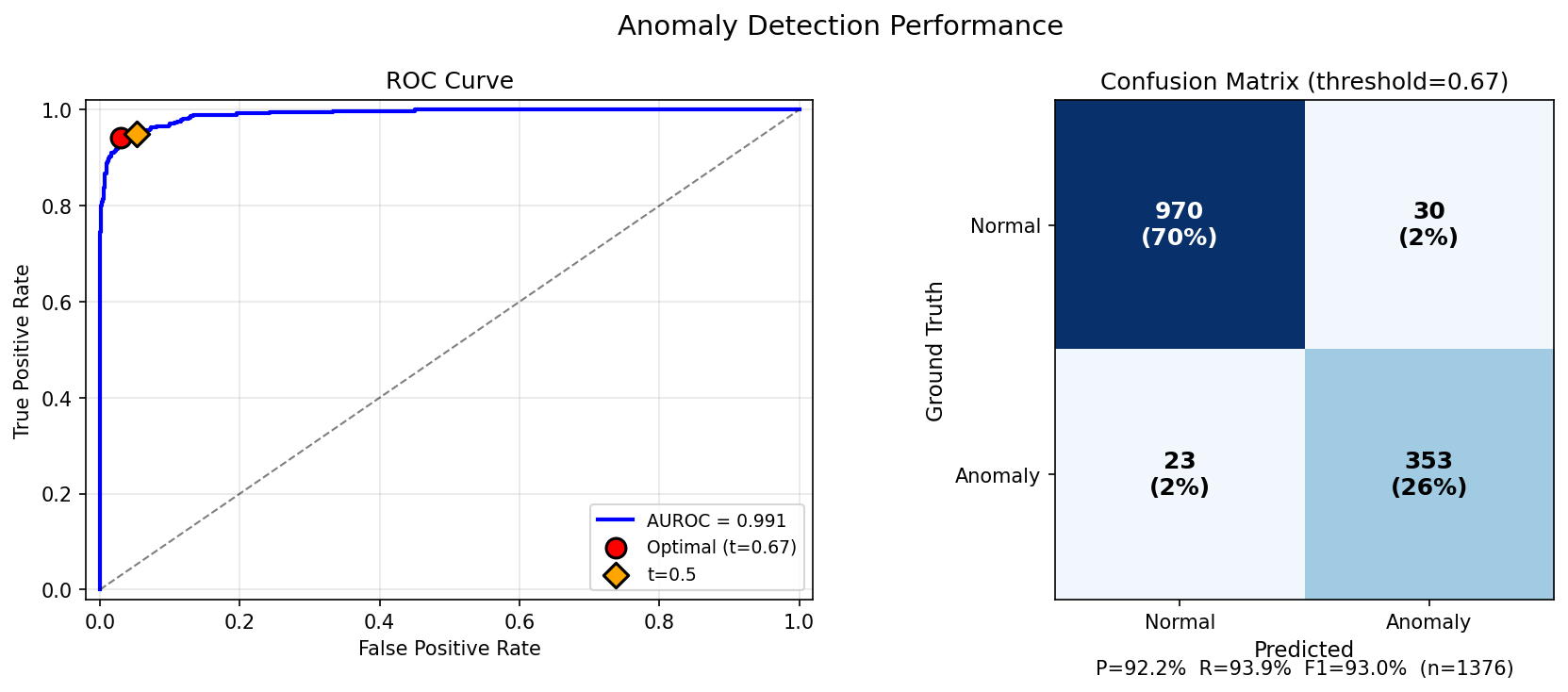}
    \caption{Overall anomaly detection performance on 5-minute detection window. Left: ROC curve with AUROC = 0.991 and operating points for the default and optimal thresholds. Right: confusion matrix at the optimal threshold ($t=0.67$).}
    \label{fig:roc_confusion}
\end{figure}

\subsubsection{Overall Detection Performance}
The embedding-based anomaly detector displays strong discriminative performance. Operating at a 5-minute window, the receiver operating characteristic yields an AUROC of $0.991$ (Figure~\ref{fig:roc_confusion}). Utilizing an optimal threshold of $t=0.67$, the detector achieves 92.2\% precision, 93.9\% recall, and a 93.0\% F1 score across 1,376 evaluation windows. It triggers only 30 false positives and misses a mere 23 events. These margins show that embedding deviations provide a reliable mathematical signal for isolating anomalous behavior from standard traffic flow.


\begin{figure}
    \centering

    \begin{minipage}[t]{0.48\textwidth}
        \centering
        \includegraphics[width=\linewidth]{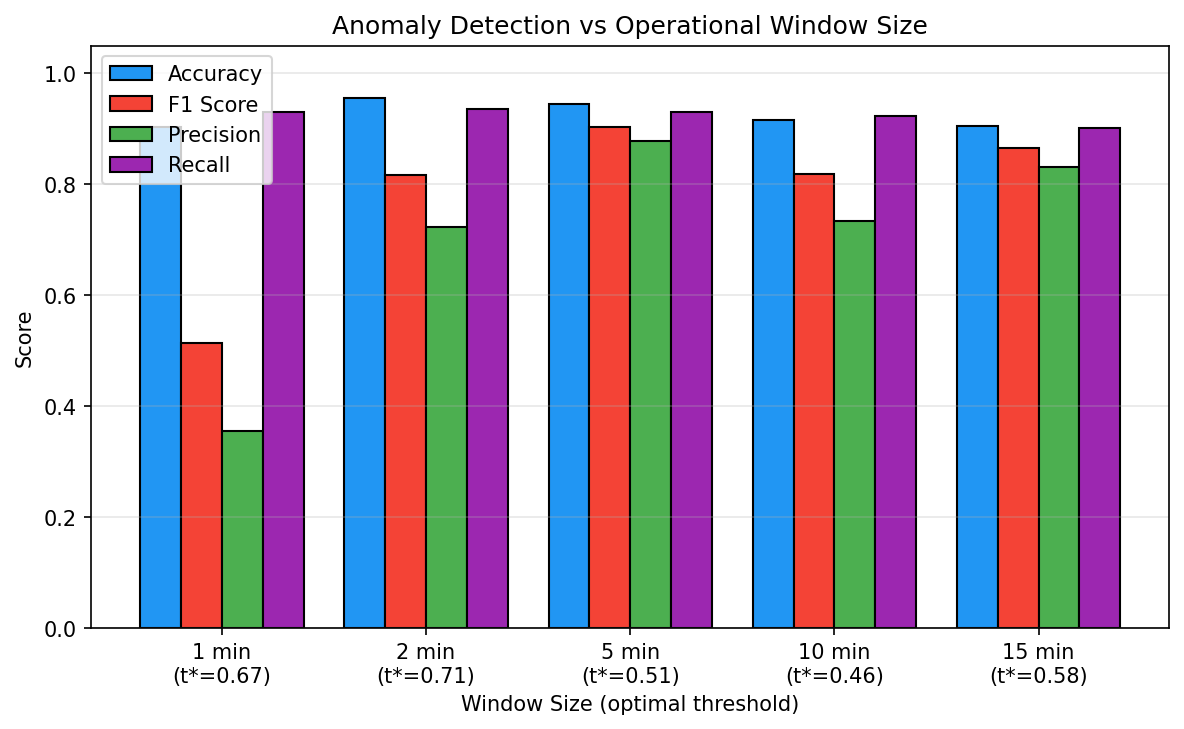}\par
        \vspace{1mm}
        \textbf{(a) Window size comparison}\par
    \end{minipage}
    \hfill
    \begin{minipage}[t]{0.48\textwidth}
        \centering
        \includegraphics[width=\linewidth]{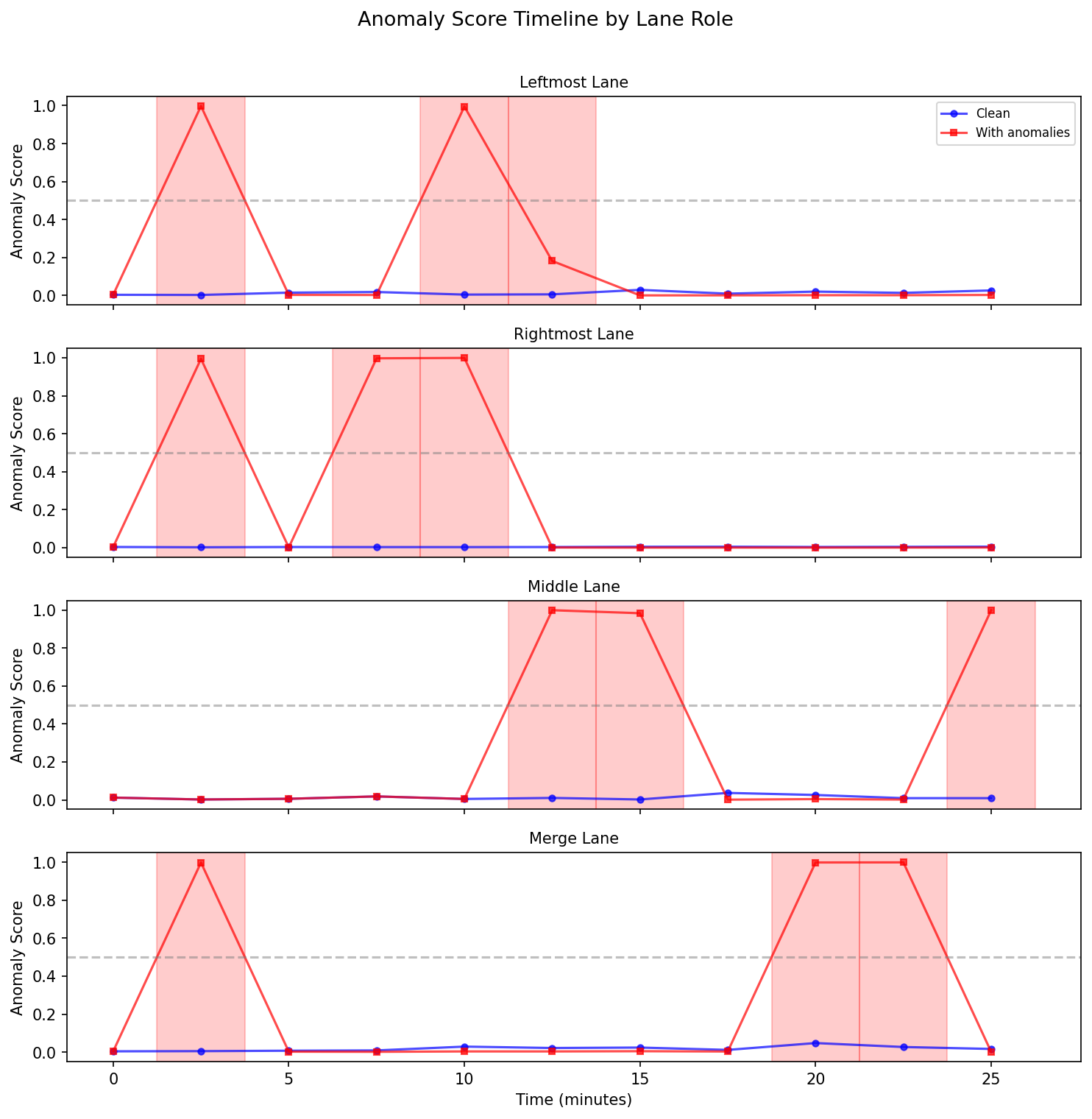}\par
        \vspace{1mm}
        \textbf{(b) Anomaly score timelines}\par
    \end{minipage}

    \caption{
    Anomaly detection evaluation.
    (a) Anomaly detection performance across operational window sizes.
    (b) Anomaly score timelines across four representative lane roles. Each panel plots the detector's predicted anomaly probability on the clean window sequence and on the same sequence with injected corruptions. Shaded bands mark windows where a synthetic anomaly was injected into the trajectory input.
    }
    \label{fig:anomaly_results}
\end{figure}

\subsubsection{Effect of Operational Window Size}
Detection fidelity intrinsically depends on the volume of temporal context available. However, performance stabilizes rapidly (Figure~\ref{fig:anomaly_results}(a)). At a severe 1-minute restriction, precision drops to 0.36, and the F1 score falls to 0.51, while recall remains high at 0.93. The detector still catches most injected anomalies, but the per-window trajectory statistics are averaged over too few samples to separate corrupted windows from noisy clean ones, driving false positives up. This lack of history prevents the formation of a stable lane-level baseline. Expanding to a 2-minute window triggers a substantial recovery (F1 $\approx$ 0.82). The detector finds its optimal operational balance at the 5-minute mark (F1 $\approx$ 0.90, precision $\approx$ 0.88, recall $\approx$ 0.93). Performance remains consistent through 10- and 15-minute intervals. Consequently, a 2-to-5-minute contextual horizon provides maximum anomaly resolution while remaining highly practical for near-real-time digital twin monitoring.

\subsubsection{Temporal Localization Across Lane Roles}
Figure~\ref{fig:anomaly_results}(b) maps anomaly score trajectories across standard leftmost, rightmost, middle, and complex merge lanes. Each panel overlays the predicted probability on the clean sequence (blue baseline) against the corrupted sequence (red), with shaded bands marking the injected windows. Under nominal conditions, these scores stay near zero. This proves the embedding holds temporal stability during standard operations. Upon anomaly injection, the scores spike sharply above the $0.5$ threshold inside the injected windows, with magnitude varying by anomaly type. They return cleanly to baseline the moment the event concludes. Crucially, this reactive behavior is uniform across all structural lane roles. The anomaly signal is not biased toward a specific lane archetype.

\subsection{Behavior-Conditioned Geometry Generation}
The final experiment evaluates whether the learned representation can physically guide the generation of new lane geometries that adhere to targeted semantic specifications. The generator receives a target specification and a spatial anchor and outputs candidate centerlines, which are re-encoded and scored for semantic compliance.

\begin{figure*}
    \centering
    \begin{minipage}[t]{\linewidth}
        \centering
        \includegraphics[width=\linewidth]{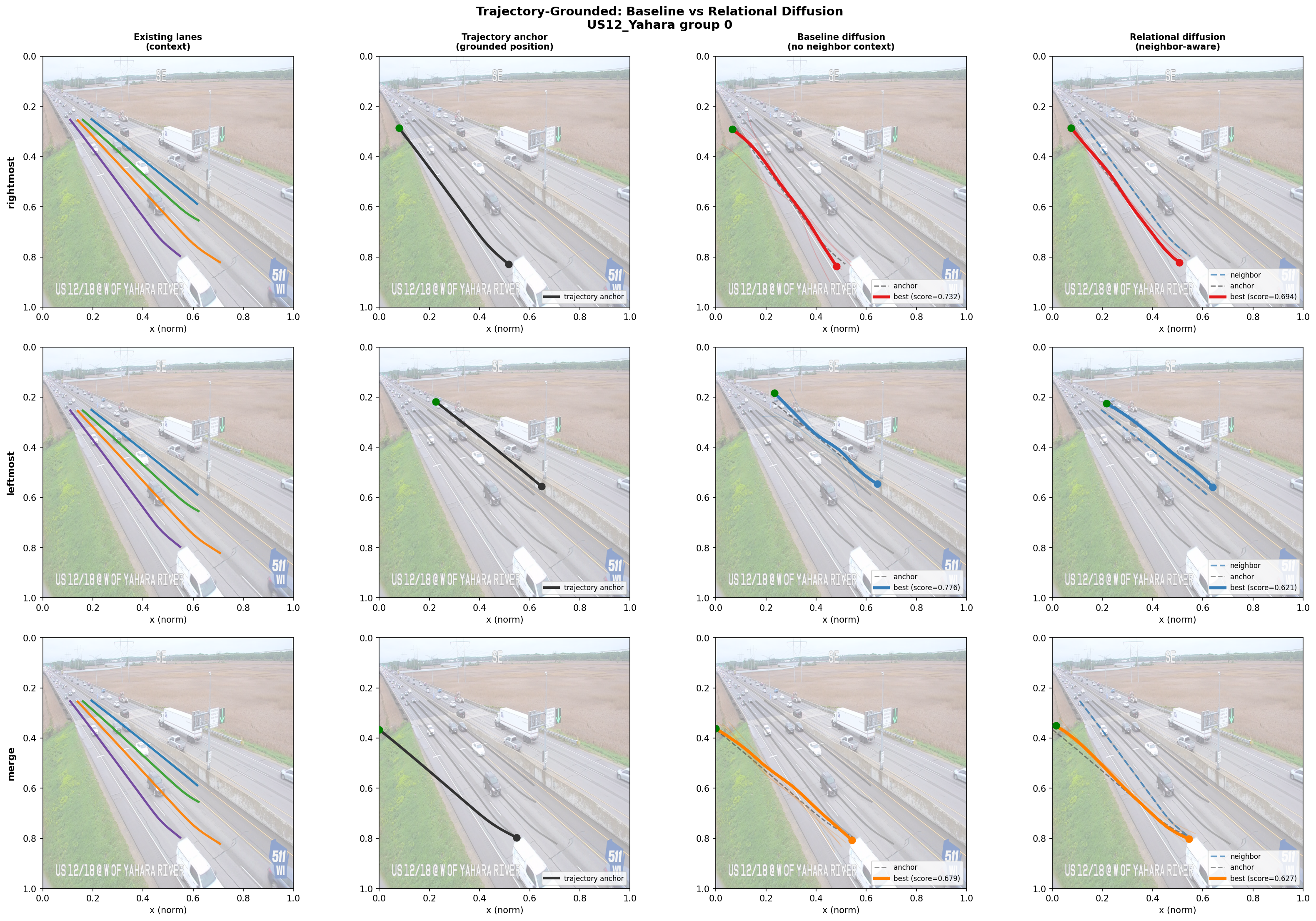}
    \end{minipage}
    \caption{Qualitative examples of behavior-conditioned lane generation at \texttt{US12\_Yahara} (group~0), comparing independent and relational conditioning for rightmost, leftmost, and merge lane generation.}
    \label{fig:generation_qualitative}
\end{figure*}

\subsubsection{Spec-Conditioned and Relational Generation}
Figure~\ref{fig:generation_qualitative} showcases generation performance at the \texttt{US12\_Yahara} site. The module produces plausible centerlines for rightmost, leftmost, and merge roles when conditioned on lateral role and anchor alone. For standard (edge) lanes, both independent and relational variants yield geometries that visually align with the real lane structure, with rightmost and leftmost candidates exhibiting the lowest semantic reconstruction errors. This confirms the generated lines remain faithful to the target embedding.

Relational conditioning explicitly exposes the generator to the neighboring-lane context. This constrains the output toward locally consistent spacing. The generated geometries track the surrounding lane bundle rather than drifting independently, as is visible in the no-neighbor variant.

\subsubsection{Generation Quality Across Sites}

\begin{table}[htbp]
\centering
\caption{Quantitative evaluation of behavior-conditioned lane generation (38 lane groups, 570 candidates, 5 candidates per spec per group). FGD and curvature smoothness are reported in both raw and filtered forms; the filtered forms exclude candidates flagged by the curvature-outlier criterion.}
\label{tab:generation_metrics}
\begin{tabular}{@{}llc@{}}
\toprule
\textbf{Category} & \textbf{Metric} & \textbf{Value} \\
\midrule
\textit{Specification Accuracy} & Rightmost Lane & 98.4\% \\
                             & Leftmost Lane & 99.5\% \\
                             & Merge Lane & 65.8\% \\
                             & \textbf{Overall Accuracy} & \textbf{87.9\%} \\
\midrule
\textit{Per-spec Chamfer} & Rightmost & 0.039 \\
                          & Leftmost & 0.040 \\
                          & Merge    & 0.052 \\
\midrule
\textit{Geometric \& Spatial} & Candidate Diversity (Mean L2) & 0.1737 \\
                              & Spatial Coherence & 0.1003 \\
                              & Chamfer Distance (overall) & 0.0450 \\
                              & Fréchet Geometry Distance (raw / filt.) & 0.0179/0.0254 \\
                              & Curvature Smoothness (raw / filt.) & 2.727/0.0220 \\
                              & Smoothness Outliers & 7.0\% (38/570) \\
\bottomrule
\end{tabular}
\end{table}

Quantitative evaluation across 38 camera groups and 570 generated candidates (Table~\ref{tab:generation_metrics}) supports these observations. The generator attains 99.5\% lateral specification accuracy for leftmost lanes and 98.4\% for rightmost lanes, with an overall accuracy of 87.9\%. Merge lanes account for most of the residual error (65.8\% accuracy), examining where merge-specified candidates actually land shows that failures do not scatter randomly but bifurcate into leftmost or rightmost geometries, suggesting that the model has learned ``merge'' as a region of interpolation between the two edge-lane modes rather than as a distinct topological class. Per-spec Chamfer distances tell the same story: merge candidates sit at $0.052$, roughly $30\%$ higher than the $0.039$--$0.040$ seen on rightmost/leftmost, so the gap is not only categorical but also geometric.

Aggregate geometric fidelity is strong. A mean Chamfer distance of $0.0450$ indicates close pointwise alignment with the real roadway geometry. Fr\'echet Geometry Distance (FGD) is $0.0179$ on the full candidate set and $0.0254$ after excluding candidates flagged as curvature outliers; in both regimes, the distributional shape of the generated lanes closely matches the reference. The generator avoids representation collapse: mean pairwise $L_2$ distance is $0.1737$, so the system produces a diverse ensemble of candidates rather than repeating a single solution. Finally, the curvature-smoothness analysis flags $7.0\%$ (38 / 570) of candidates as rough outliers -- high variance in heading-change angles along the polyline -- leaving $93\%$ whose curvature profiles fall within the smooth range defined by the threshold.

\section{Conclusions}

This research introduces \textbf{GeoLaneRep}, a behavior-grounded representation framework that enhances traffic digital twins by mapping static lane geometry, observed vehicle trajectories, and operational descriptors into a single shared embedding. Trained jointly with contrastive cross-camera alignment, auxiliary role supervision, and temporal anomaly detection, the encoder achieves a $0.004$ lateral-rank error and an edge-role F1 of $1.000$ under a leave-one-camera-out protocol, and an AUROC of $0.991$ on per-window anomaly detection. Consequently, it produces a highly transferable embedding that generalizes seamlessly across heterogeneous camera deployments. Furthermore, a FiLM-conditioned diffusion module uses these embeddings to synthesize candidate lane geometries that satisfy targeted operational specifications at $87.9\%$ overall accuracy. The same encoder weights drive all three tasks, supporting the central claim that a single behavior-aware representation can serve heterogeneous lane-level digital-twin applications.

The evaluation also surfaces an asymmetry between representation and generation quality. The encoder produces a stable, highly discriminative embedding space, but the generated lane geometry trails this fidelity, particularly for merge specifications ($65.8\%$ accuracy versus $98$--$99\%$ for edge lanes). This gap likely reflects compounded upstream uncertainty rather than a generator-side defect alone: imperfect trajectory extraction, occlusion, lane-marking ambiguity, and limited coverage of rare lane types each weaken the geometric supervision the diffusion module receives. A second limitation is that the generation scheme operates lane-wise and does not yet model the mutual dependence among neighboring lanes that real roadway geometry imposes. Third, the framework prioritizes semantic consistency over engineering-grade geometric fidelity. This allows training on noisy roadside observations without curated HD-map labels, but limits direct deployability for tasks that need vertical alignment, curvature continuity, or precise lane-boundary reconstruction. 

Looking forward, closing the representation-generation gap calls for richer geometric supervision: multi-view reconstruction, depth estimation, LiDAR fusion, or BIM-derived lane references would give the diffusion module a cleaner target without disturbing the behavioral embedding. Moving from lane-wise to corridor-level synthesis would require coupled multi-lane generation under topological and geometric constraints. For example, graph-structured decoders over lane groups, or relational denoisers conditioned on the neighboring-lane embeddings already produced by the cross-lane attention module. A third direction is interfacing the embedding with connected vehicle technologies. For example, vehicle-to-everything messages provide a machine-readable lane topology that could either supervise the representation or, conversely, be inferred or repaired by it when annotations are incomplete. Together, these extensions would carry GeoLaneRep beyond lane-level interpretation toward the coordinated, geometrically faithful interventions that constructive digital twins ultimately require.

\bibliographystyle{cas-model2-names}
\bibliography{references}

\end{document}